\documentclass[11pt]{article}
\usepackage[preprint,nonatbib]{neurips_2025}

\usepackage[numbers,sort&compress]{natbib}
\usepackage[utf8]{inputenc} 
\usepackage[T1]{fontenc}
\usepackage{hyperref}       %
\usepackage{url}    
\usepackage{amssymb}
\usepackage{float}
\usepackage{amsmath}
\usepackage{graphicx}
\usepackage{enumitem}
\usepackage{cleveref}
\usepackage{algorithm}     
\usepackage{algorithmic}
\usepackage{tabularx}   
\usepackage[labelfont=bf, labelsep=period]{caption}
\usepackage{booktabs}
\usepackage{amsthm}  

\Crefname{figure}{Fig.}{Figs.} 
\newtheorem{definition}{Definition}

\newcommand{\afftextA}{Department of Computer Science and Artificial Intelligence, Andalusian Research Institute in Data Science and Computational Intelligence (DaSCI), University of Granada, Spain}
\newcommand{\afftextB}{Department of Software Engineering, Andalusian Research Institute in Data Science and Computational Intelligence (DaSCI), University of Granada, Spain}
\newcommand{\afftextC}{School of Management, Hangzhou Dianzi University, China}
\newcommand{\afftextD}{Shaanxi Key Laboratory of Information Communication Network and Security, Xi’an University of Posts \& Telecommunications, China}
\newcommand{\corrstar}{\textsuperscript{*}} 

\title{Addressing Data Quality Decompensation in Federated Learning via Dynamic Client Selection}

\author{ 
	Qinjun Fei \textsuperscript{a}
    \And
    Nuria Rodríguez-Barroso \textsuperscript{a,}  \corrstar
    \And
    María Victoria Luzón \textsuperscript{b}
    \And
    Zhongliang Zhang \textsuperscript{c, d}
    \And
    Francisco Herrera \textsuperscript{a}
    \And
    \\[\medskipamount]
    \hfill
    \parbox{\textwidth}{\centering\small
    \textsuperscript{a} \textit{\afftextA}\\[0.5ex]
    \textsuperscript{b} \textit{\afftextB}\\[0.5ex]
    \textsuperscript{c} \textit{\afftextC}\\[0.5ex]
    \textsuperscript{d} \textit{\afftextD}
    }
    \hfill
}

\begin{document}

\maketitle
\renewcommand\thefootnote{\fnsymbol{footnote}}
\footnotetext[1]{Corresponding Author: Nuria Rodríguez-Barroso}
\footnotetext[0]{Email: \texttt{qinjun@correo.ugr.es} (Qinjun Fei), \texttt{rbnuria@ugr.es} (Nuria Rodríguez-Barroso), \texttt{luzon@ugr.es} (María Victoria Luzón), \texttt{zzl19860210@126.com} (Zhongliang Zhang), \texttt{herrera@decsai.ugr.es} (Francisco Herrera)}
\renewcommand\thefootnote{\arabic{footnote}}

\begin{abstract}
In cross-silo Federated Learning (FL), client selection is critical to ensure high model performance, yet it remains challenging due to data quality decompensation, budget constraints, and incentive compatibility. As training progresses, these factors exacerbate client heterogeneity and degrade global performance. Most existing approaches treat these challenges in isolation, making jointly optimizing multiple factors difficult. To address this, we propose Shapley-Bid Reputation Optimized Federated Learning (SBRO-FL), a unified framework integrating dynamic bidding, reputation modeling, and cost-aware selection. Clients submit bids based on their perceived data quality, and their contributions are evaluated using Shapley values to quantify their marginal impact on the global model. A reputation system, inspired by prospect theory, captures historical performance while penalizing inconsistency. The client selection problem is formulated as a 0-1 integer program that maximizes reputation-weighted utility under budget constraints. Experiments on FashionMNIST, EMNIST, CIFAR-10, and SVHN datasets show that SBRO-FL improves accuracy, convergence speed, and robustness, even in adversarial and low-bid interference scenarios. Our results highlight the importance of balancing data reliability, incentive compatibility, and cost efficiency to enable scalable and trustworthy FL deployments.
\end{abstract}

\noindent\textbf{Keywords:} Federated learning, Reputation-based client selection, Data quality decompensation, Incentive mechanism, Client selection, Shapley value

\section{Introduction}\label{introduction}
Federated Learning (FL) has emerged as a distributed machine learning paradigm that enables collaborative model training without centralizing data, effectively mitigating privacy concerns and reducing communication overhead \citep{konevcny2016federated,mcmahan2017communication}. FL initially gained attention in large-scale cross-device settings, such as mobile networks and edge computing, and has since been widely adopted in decentralized AI applications \citep{badr2023privacy,bonawitz2019towards}. Meanwhile, cross-silo FL collaboration continues to evolve, introducing additional complexities beyond those encountered in cross-device settings. These challenges, including data quality decompensation, budget constraints, and incentive mechanisms, create a dynamic landscape of interacting practical factors, making cross-silo scenarios particularly demanding \citep{huang2022cross,zhang2024survey}.

Unlike cross-device FL, which involves a large number of resource-constrained devices, cross-silo FL typically involves a limited number of organizational clients such as hospitals, financial institutions, and industrial enterprises that have abundant data and computational resources \citep{rahman2023federated}. However, the quality of data from these sources can vary widely. For instance, in healthcare-oriented FL, a hospital may inadvertently provide mislabeled patient records due to inconsistent annotation practices or human error, while in industrial IoT applications, sensor calibration mismatches across factories can inject systematic biases into the global model. Moreover, adversarial behaviors like label flipping or model poisoning further distort the training process, increasing the risk of unreliable global updates \citep{li2020review, rodriguez2023survey, sun2021data}. This variability in data quality introduces a persistent issue referred to as data quality decompensation, where the accumulation of unreliable client updates over multiple rounds leads to global model instability and performance degradation, despite individual datasets remaining unchanged. Consequently, robust mechanisms are required to assess and selectively filter clients, ensuring that only high-quality contributions drive model optimization.

Beyond client selection, another key challenge is incentivizing high-quality clients to participate consistently \citep{qi2022high}. The value of data is highly task-dependent and as data become increasingly commoditized, clients expect fair compensation \citep{yuan2024decentralized}. Furthermore, local training incurs resource costs, including computation, storage, and communication, which require financial incentives to sustain engagement \citep{wen2023survey}. However, in real-world FL systems with financial incentives, the FL initiator faces budget constraints, limiting client compensation. This requires careful budget allocation across data valuation, incentive mechanisms, and cost-effective client selection.

Most client selection strategies in FL primarily focus on optimizing a single aspect, such as maximizing convergence speed \citep{lee2022data}, designing incentive mechanisms \citep{wang2023incentive}, or improving communication efficiency \citep{panigrahi2023feddcs}. Although effective in their respective domains, these approaches often overlook the interdependencies between data quality decompensation, incentive compatibility, and budget constraints. In practical cross-silo scenarios, a unified framework that coherently integrates these dimensions is essential for jointly optimizing data reliability, cost-effectiveness, and equitable participation incentives.

To address these challenges, we hypothesize that integrating a bidding mechanism with a dynamic reputation system can mitigate data quality decompensation while ensuring budget efficiency. However, historical reputation alone may not fully capture the reliability of clients, as clients with similar scores can exhibit varying stability. To account for this, selection should incorporate risk sensitivity, prioritizing clients with consistent contributions and cost-effectiveness. This approach is expected to improve the stability of the model and optimize the participation incentives in cross-silo FL.

This paper proposes Shapley-Bid Reputation Optimized Client Selection for Federated Learning (SBRO-FL), a method that integrates reputation-driven bidding with cost-aware optimization. Before training, clients submit bid prices based on perceived data quality, aligned with budget constraints set by the FL initiator. SBRO-FL maintains a reputation value to track historical contributions, while reputation scores, adjusted through prospect theory, guide selection by accounting for risk sensitivity. The selection process is formulated as a 0-1 integer programming problem, optimizing reputation-weighted utility under budget constraints. At the end of each round, the Shapley value quantifies marginal contributions, shaping a cost-effectiveness metric that updates reputation values and informs future selection. A dynamic penalty mechanism discourages unreliable participation, reinforcing system stability. By incorporating bidding incentives and dynamic reputation tracking, the proposed method increases FL efficiency and encourages stable client participation. Specifically, we make the following contributions:
\begin{enumerate}[label=(\arabic*)]
    \item \textbf{Unified Client Selection Framework:} Unlike existing approaches that optimize a single aspect, SBRO-FL jointly optimizes data quality decompensation, incentive compatibility, and budget constraints in a unified selection framework, enhancing practical applicability in cross-silo FL.
    \item \textbf{Reputation-Driven Bidding:} A novel bidding mechanism is introduced, where clients propose bid prices based on perceived data quality. This approach provides a practical incentive alignment strategy under budget constraints, fostering stable and cost-efficient client engagement.
    \item \textbf{Shapley Value-Based Contribution Assessment:} SBRO-FL leverages the Shapley value to quantify the marginal impact of each client on the global model. This contribution metric is incorporated into a cost-effectiveness measure, which influences both reputation updates and future selection probabilities.
    \item \textbf{Budget-Constrained Optimization for FL:} The client selection problem is formulated as a 0-1 integer programming model, optimizing reputation-weighted utility while adhering to strict budget constraints. This ensures that selected clients provide an optimal balance of data reliability and participation cost.
\end{enumerate}
The experiments compared SBRO-FL with three baselines: (1) random client selection, (2) selection of all clients, and (3) random selection from a high-quality subset. These baselines represent varying selection strategies, from uncontrolled participation to partial quality-aware selection. SBRO-FL consistently outperformed these baselines in four datasets, improving the accuracy of the global model by an average of 7.14\% compared to random selection after a fixed number of rounds. Its performance closely approached that of an idealized selection strategy in which only high-quality clients were chosen, free of label noise. This highlights SBRO-FL’s ability to infer client reliability dynamically, effectively balancing data quality, incentives, and budget constraints.

\section{Background}\label{background}
In this section, we provide an overview of key concepts and related work. \Cref{FLdescription} covers the fundamentals of FL, \Cref{clientselection} reviews client selection strategies that focus on resource-oriented and performance-oriented approaches, and \Cref{incentivemechanisms} explores incentive mechanisms aimed at maintaining client participation in FL.

\subsection{Federated Learning}\label{FLdescription}
\begin{figure*}[t]
\centering
\includegraphics[width=0.7\textwidth]{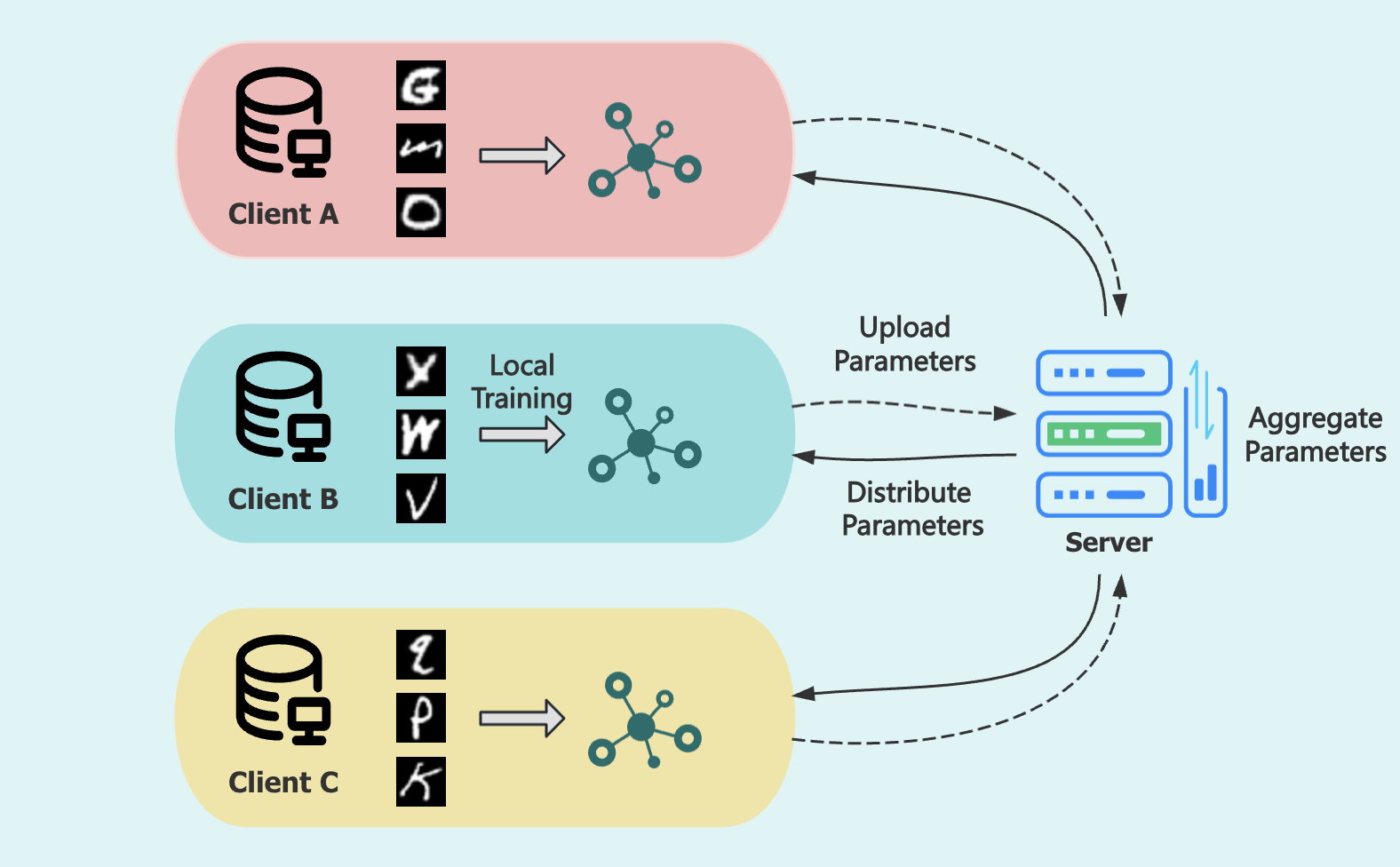}
\caption{Federated learning example: client–server architecture.}
\label{flcs}
\end{figure*}

FL is a distributed machine learning paradigm designed to protect data privacy and has drawn considerable attention \citep{mcmahan2017communication}. In FL, each client receives the global model at the start of each round, trains it locally on private data, and then sends the updated parameters for aggregation into a new global model, as illustrated in \Cref{flcs}. This process repeats over multiple rounds until a stopping condition is met. FL enables collaborative model training across decentralized data sources without requiring data sharing, making it ideal for privacy-sensitive applications.

A typical FL architecture consists of a central server and multiple clients, denoted as $\mathcal{C} = \{ C_1, \ldots, C_n \}$. The goal of FL is to collaboratively train a global model $\mathbf{w}$ by leveraging decentralized data from multiple clients. This is achieved by iteratively solving the following optimization problem:
\begin{equation}
\min_{\mathbf{w}} \sum_{i=1}^{n} \frac{|D_i|}{\sum_{j=1}^{n} |D_j|} \ell(\mathbf{w}; D_i)
\end{equation}
where \(\mathbf{w}\) denotes the global model parameters, \(D_i\) represents the local dataset of client \(C_i\), with \(|D_i|\) denoting its number of data samples, serving as the weighting factor in the aggregation process. The term \(\ell(\mathbf{w}; D_i)\) represents the local loss function evaluated on \(D_i\).
One of the most widely used algorithms in FL is FedAvg, which iteratively solves the optimization problem. At the beginning of each round $t$, the global model parameters $\mathbf{w}^t$ are sent to all participating clients. Each client independently updates the global model based on its local data $D_i$ by performing several steps of a local optimization algorithm on the corresponding loss function:
\begin{equation}
\mathbf{w}_i^t = \mathbf{w}^t - \eta \mathbf{g}_i^t
\end{equation}
where \(\eta\) is the learning rate and $\mathbf{g}_i^t$ represents the gradient or update direction calculated on the local data. After local training, the clients transmit their updated parameters \(\mathbf{w}_i^t\) back to the server. The server then aggregates these updates by computing a weighted average, based on the size of each client’s dataset, to yield the new global model parameters for the next round:
\begin{equation}
\mathbf{w}^{t+1} = \sum_{i=1}^{n} \frac{|D_i|}{\sum_{j=1}^{n} |D_j|} \mathbf{w}_i^t.
\end{equation}
This process continues iteratively until a convergence criterion is satisfied, such as achieving a specified number of rounds or achieving a target accuracy.

\subsection{Client Selection}\label{clientselection}
In the context of FL, effective client selection is essential to improve model performance and optimize resource utilization. The distinction between resource-oriented and performance-oriented selection strategies clarifies their respective impacts on training efficiency and model reliability.

Resource-oriented criteria emphasize the relevance of the computational capabilities of clients and the efficiency of communication, enabling faster, more reliable, and resource-efficient processes. The FedCS framework proposed by \citet{nishio2019client} prioritizes clients capable of completing a training round within a predefined deadline. The selection process analyzes the update efficiency of each client, considering their computational capabilities and the conditions of the wireless channel. From another perspective, this approach allows FL to involve more clients in training in a shorter time, thereby accelerating the overall FL process. \citet{yu2021jointly} designed the energy and latency-aware resource management and client selection algorithm (ELASTIC), which not only aims to maximize the number of selected clients, but also minimizes the total energy consumption based on the CPU frequency and transmission power of the clients. These approaches are particularly effective in FL between devices, where resource limitations are more pronounced \citep{mayhoub2024review}. 

Performance-oriented criteria focus primarily on data quality and model attributes, in order to select clients that significantly improve the effectiveness of the global model. Given the substantial influence of training data volume on model performance, some approaches prioritize clients based on the quantity of their data. For example, \citet{jeon2020optimal} proposed a selection strategy that considers the quantity of client data, the quality of communication, and the residual energy. Similarly, \citet{li2019convergence} introduced an unbiased sampling scheme based on a polynomial distribution, where the probability of selection of a client is proportional to their quantity of data. However, the relationship between data quantity and global model quality is often non-linear and complex, particularly in scenarios with inconsistent data quality and Non-IID (independent and identically distributed) data distributions. To address these complexities, \citet{fraboni2021clustered} compared two clustering-based sampling methods: one based on sample size and the other on model similarity. Their findings demonstrated that the model similarity-based method improved convergence speed and stability, especially as data heterogeneity increased. Beyond data quantity, other important indicators of client data quality include data distribution \citep{lee2022data}, local model parameters \citep{balakrishnan2021diverse,zhao2022participant}, and local model loss \citep{cho2020client}. These approaches are more prevalent in cross-silo FL, where the focus is on optimizing global model performance.

While existing methods optimize either computational efficiency or model performance, they typically overlook economic constraints. In cross-silo FL, client selection must also account for budget limitations, where high-quality data alone does not guarantee selection if participation costs are excessive. This necessitates a balance between data reliability and financial feasibility. Moreover, since incentives directly impact client engagement, selection strategies must align compensation with contribution quality.

\subsection{Incentive Mechanisms}\label{incentivemechanisms}
In FL, the selection of clients presupposes the availability of a pool of candidates willing to contribute. Clients invest significant resources, including data and computational power, but without fair incentives, many may opt out \citep{khan2020federated}. This highlights the need for robust incentive mechanisms to maintain engagement, especially when limited budgets restrict the selection of all clients.

Reputation has emerged as a key incentive mechanism in client selection, directly related to both data quality and client reliability \citep{mayhoub2024review, chen2024credible}. Several works have proposed using reputation as a benchmark for assessing client trustworthiness and performance. For example, \citet{kang2019incentive} designed a reputation-based client selection framework where clients’ reliability is measured using subjective logic models. Similarly, \citet{song2021reputation} adopted a beta distribution-based reputation model to systematically assess client trustworthiness. While these frameworks enhance selection reliability, they often overlook how reputation interacts with budget constraints to shape sustainable incentive strategies. In practice, aligning reputation-based selection with economic feasibility is essential for maintaining long-term engagement and cost-effective participation.

Moreover, auction-based mechanisms have gained prominence in FL to align client incentives with the server’s objectives. \citet{pang2022incentive} designed an incentive mechanism using a procurement auction model and a greedy algorithm to optimize client selection and scheduling, aiming to minimize social cost by efficiently distributing training participation across iterations.  \citet{jiao2020toward} proposed a budget optimization framework for multi-session FL auctions, leveraging hierarchical reinforcement learning to refine strategic pricing adjustments by data consumers. In parallel, game-theoretic models, particularly Stackelberg games, have also been employed to balance incentives and resource allocation. \citet{sarikaya2019motivating} and \citet{khan2020federated} formulated Stackelberg-based strategies where the FL server acts as a leader setting rewards, while clients respond by adjusting their participation efforts. However, many of these methods operate under the assumption that data quality is relatively uniform or fail to explicitly address the heterogeneous nature of cross-silo data sources, which may also be prone to adversarial behavior.

While existing methods have separately addressed data-driven selection, incentive mechanisms, and budget-aware strategies, their independent treatment often overlooks the interdependencies among these factors. To address these interrelated challenges, our approach jointly considers data quality decompensation, incentive compatibility, and budget constraints, thereby aligning client participation with both model reliability and economic feasibility. The following sections detail the methodology.

\section{Shapley-Bid Reputation Optimized Client Selection}\label{methodology}
SBRO-FL integrates reputation-driven bidding with cost-aware optimization to improve efficiency in FL under budget constraints. The framework dynamically evaluates client reliability and cost-effectiveness, adapting selection strategies to evolving training conditions. For clarity, \Cref{workflow} illustrates its workflow within a traditional FL setup.
\subsection{Workflow of SBRO-FL}\label{workflow}
\begin{figure*}[t]
\centering
\includegraphics[width=0.6\textwidth,clip]{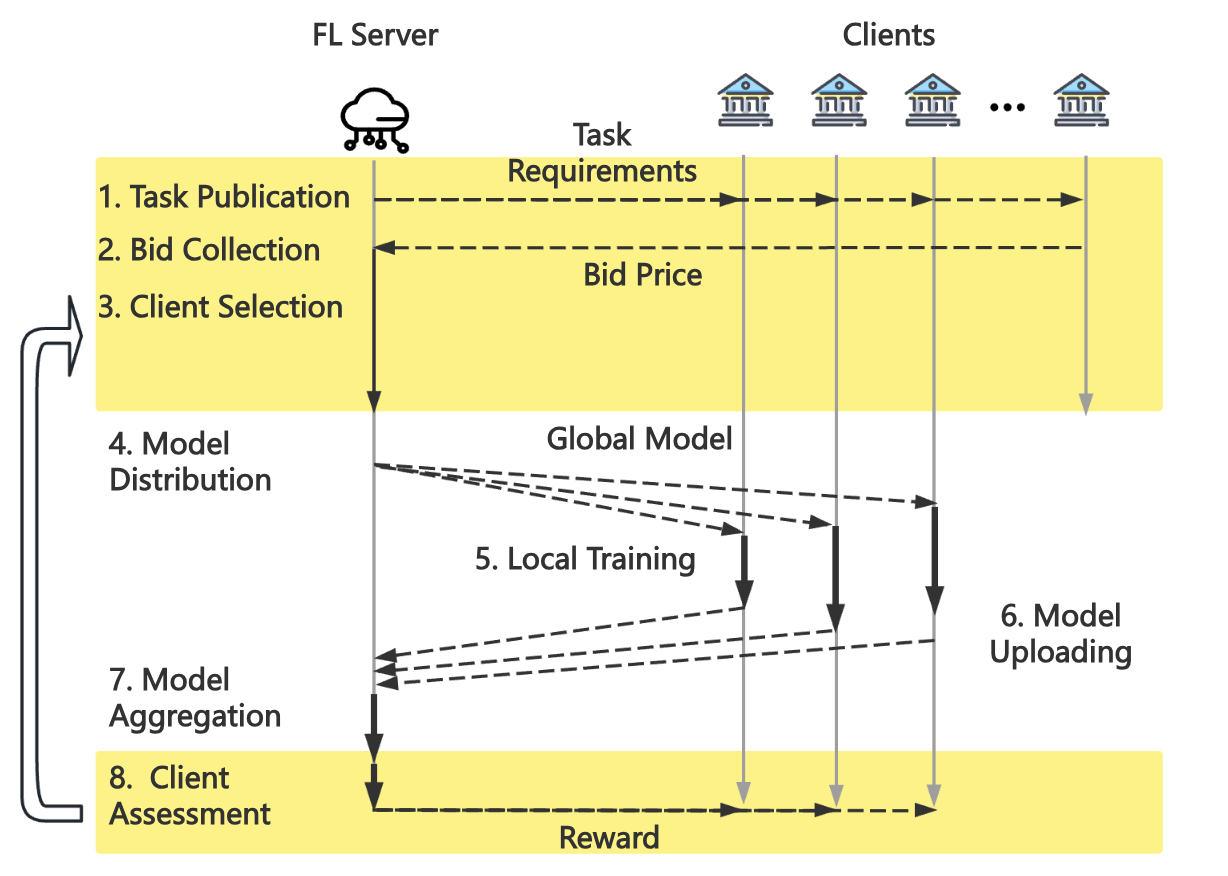}
\caption{Workflow of SBRO-FL within the traditional FL framework.}
\label{fig:mylabel1}
\end{figure*}
\Cref{fig:mylabel1} illustrates the workflow of this method. The steps highlighted in yellow indicate the key differences from standard FL procedures. The communication flow for each round is as follows:
\begin{enumerate}
    \item \textbf{Task Publication:} The central server publishes the task requirements, including the data type, model size, and training methodology, to all potential clients.
    \item \textbf{Bid Collection:} Clients submit bid prices based on their estimated capacity to meet the task requirements.
    \item \textbf{Client Selection:} The server calculates a reputation score for each client based on historical performance and selects clients through an optimization model formulated as a 0-1 integer programming problem.
    \item \textbf{Model Distribution:} The central server distributes the specification of the FL task, including the type and parameters of the global model.
    \item \textbf{Local Training:} The selected clients train the model locally using their private datasets.
    \item \textbf{Model Uploading:} The selected clients upload the locally trained model parameters to the server.
    \item \textbf{Model Aggregation:} The central server aggregates all the local model parameters to update the global model for the current communication round.
    \item \textbf{Client Assessment:} After allocating rewards based on their bid prices, the central server calculates the selected clients’ contributions to the global model using the Shapley value and updates their reputations by considering both their contributions and submitted bids.
\end{enumerate}
The following subsections follow the natural execution order of the client selection process in FL. Since the selection model aims to maximize reputation-based utility, \Cref{reputationvaluation} first introduces the reputation score, which serves as the foundation for the clients selections model in \Cref{selectionmodel}. To reflect the dynamic nature of the overall system, the client assessment phase at the end of each training round employs a Shapley value-based evaluation method in \Cref{evaluationclient} to quantify client contributions, which also serves as an indicator of data quality. This assessment then forms the basis for reputation updates in \Cref{reputationupdate}, ensuring an adaptive selection process over time.

\subsection{Reputation Values and Scores}\label{reputationvaluation}
In FL, selecting reliable clients over multiple training rounds is crucial for mitigating the risks posed by poor-quality data. To achieve this, we introduce a reputation system that dynamically evaluates and tracks client behavior across rounds, ensuring that clients with a consistent history of meaningful contributions are prioritized in future selections.

Each client is assigned a reputation value, denoted as $\mathcal{R} = \{R_1, \dots, R_n\}$ for clients $\mathcal{C} = \{C_1, \dots, C_n\}$. This reputation value serves as a measure of long-term reliability and is continuously updated after each training round. Initially, each client's reputation is set to zero and evolves based on their observed contributions. The details of the update procedure, including its influencing factors and process, are provided in \Cref{evaluationclient} and \Cref{reputationupdate}.

Although $R_i$ provides a long-term measure of client reliability, directly using it for selection may overlook the disproportionate impact of data quality decompensation. Specifically, poor-quality contributions tend to degrade the global model more significantly than high-quality updates improve it \citep{fang2020local}. 

\begin{figure*}[t]
    \centering
    \includegraphics[width=0.4\textwidth]{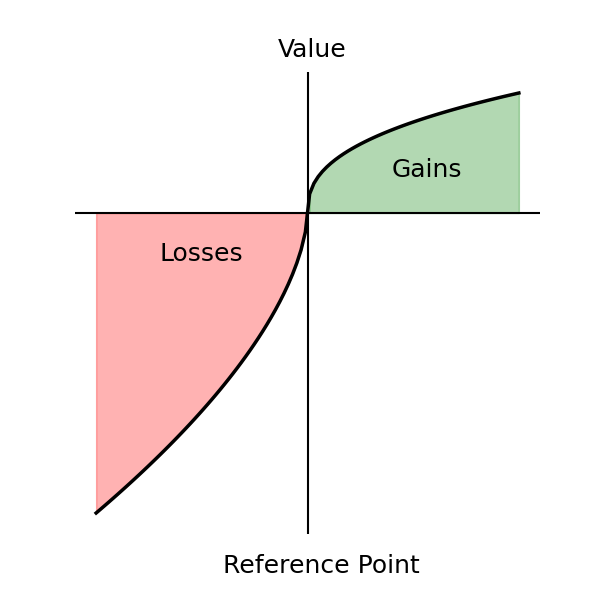}
    \caption{Prospect theory value function. The X-axis represents gains and losses relative to a reference point, while the Y-axis represents perceived value. The function is asymmetric: losses have a steeper curve than gains, reflecting loss aversion, meaning individuals perceive losses more strongly than equivalent gains. Conversely, gains exhibit diminishing sensitivity, meaning the perceived impact of additional gains decreases as they accumulate.}
    \label{fig:prospect_theory_curve}
\end{figure*}

To address this, we apply a transformation process inspired by prospect theory \citep{kahneman2013prospect}. As illustrated in \Cref{fig:prospect_theory_curve}, the prospect theory function emphasizes loss aversion, meaning that reductions in reputation (i.e., poor contributions) have a greater impact than equivalent increases. This transformation converts reputation values into reputation scores, denoted as \( \mathcal{Z}(\mathcal{R}) = \{ z(R_1), \dots, z(R_n) \} \) in \Cref{reputationscore}, which are subsequently used in the client selection process.

\begin{definition}\label{reputationscore}
    \textit{Reputation Score Function} \( z(R_i) \). \textnormal{It computes the reputation score of client $C_i$ by applying the prospect theory value function, which accounts for loss aversion and the asymmetric treatment of low- and high-quality clients following}
\begin{equation}
\begin{aligned}
z(R_i) = 
\begin{cases} 
\gamma (R_{\text{th}} - R_i)^\beta & \text{if} \quad R_i \leq R_{\text{th}}\\
(R_i - R_{\text{th}})^\alpha &  \text{if} \quad  R_i > R_{\text{th}}
\end{cases}.
\end{aligned}
\end{equation}
\end{definition}
In practice, after each round, we compute an average reputation $R_{\text{th}} = \frac{1}{p}\sum_{i=1}^{p}R_i$ across all clients, serving as the reference point in the prospect theory function. If a client's reputation $R_i \leq R_{\text{th}}$, the function treats this as a “loss” region with heightened negativity, reflecting a higher perceived risk or potentially low-quality updates. Conversely, when $R_i > R_{\text{th}}$, the client is considered more reliable, but the gain region in the Prospect transformation grows gradually rather than steeply. This prevents excessive reliance on a few high-reputation clients, ensuring that model updates leverage diverse data sources for better generalization.

The parameters $\alpha$ and $\beta$ determine the degree of asymmetry. A higher $\beta$ leads to stronger penalties for underperforming clients, whereas $\alpha$ controls the diminishing impact of higher reputations. The specific values of $\alpha$, $\beta$, and $\gamma$ were determined through preliminary tuning and guided by common parameter choices in prospect theory literature. For instance, $\beta$ is typically set between 0.2 and 0.4 to reflect strong loss aversion, while a smaller $\alpha$ indicates faster saturation of perceived gains. Based on these considerations, we set $\alpha$ = 0.15, $\beta$ = 0.3, and $\gamma$ = 1,  as this configuration provided stable performance across multiple datasets.

The adjusted reputation score $z(R_i)$ serves as the foundation for client selection in each round, guiding the optimization process described in \Cref{selectionmodel}. By incorporating this risk-aware adjustment, our framework enhances the selection of high-quality contributors while reducing the impact of unreliable clients, ensuring a more stable and effective training process.

\subsection{Selection Model}\label{selectionmodel}
Within our reputation system, client selection must balance reliability and budget constraints to ensure effective participation. The objective is to optimize client selection by prioritizing high-reputation clients, with prospect theory-adjusted scores ensuring a risk-aware balance between reliability and budget feasibility. The client selection model is defined as follows:
\begin{definition}\label{selection_model}
\textit{Client Selection Model:} $\mathbb{S}^t = Select(\mathcal{Z}, \mathcal{B}, B_{\text{budget}}, \mathcal{H}^t)$. \textnormal{Given a set of reputation scores $\mathcal{Z} = \{z(R_1), \dots, z(R_n)\}$, bid prices $\mathcal{B} = \{B_1, \dots, B_n\}$, and a total budget $B_{\text{budget}}$, the goal is to select an optimal subset of clients while considering past selection history $\mathcal{H}^t$. The selection problem is formulated as the following integer programming model:}
\begin{equation}\label{selection_model_equation}
        \begin{aligned}
        & \max \sum_{i=1}^{n} \left(z(R_i) - z_{\text{min}}\right) \cdot \delta^{count_i} \cdot x_i \\
        & \text{s.t.} \left\{
        \begin{aligned}
        & \sum_{i=1}^{n}{B_i x_i} \le B_{\text{budget}}, \\
        & B_i \geq 0, \\
        & x_i \in \{0,1\} 
        \end{aligned}
        \right.
        \end{aligned}
        \end{equation}
\end{definition}
where:
\begin{itemize}[leftmargin=*, labelsep=0.5em, nosep]
\item The term $z_{\text{min}}$ ensures that all reputation scores are non-negative by shifting the minimum reputation score to zero. The decision variable $x_i \in \{0,1\}$ determines whether client $C_i$ is selected in round $t$. The selection history $\mathcal{H}^t = \{x_1^{t-1}, \dots, x_n^{t-1}, \dots, x_1^{t-5}, \dots, x_n^{t-5}\}$ keeps track of each client’s participation in the last five rounds. $\mathbb{S}^t = \{ C_i | x_i^t = 1 \}$ denotes the set of clients selected to participate in round $t$.
\item The adjustment factor $\delta^{count_i}$ introduces diversity in selection, where $\delta$ is a decay factor and $count_i$ is the number of times client $C_i$ has been selected in the previous five rounds:
\begin{equation}
count_i = \sum_{k=1}^{5} x_i^{t-k}.
\end{equation}
\end{itemize}

The adjustment factor $\delta^{count_i}$ mitigates over-selection bias by discouraging repeated choices in early rounds while ensuring diverse participation over time. During the early rounds, it decreases the likelihood of repeatedly selecting clients that have already been chosen multiple times, ensuring a thorough review of all clients. This allows the system to evaluate a broader range of clients, accelerating the identification of high-quality clients without relying on a narrow subset. Consequently, the model enables faster filtering based on client contributions in the initial stages. In the later rounds, the model increases the likelihood of selecting clients that have been chosen less frequently, promoting both diversity and fairness. This dynamic adjustment ensures that while high-performing clients are prioritized, valuable but underrepresented clients are not excluded for extended periods, maintaining a balanced and equitable selection process.

By integrating prospect theory-based reputation scores with a dynamic adjustment factor, this model optimizes client participation while ensuring a more adaptive and balanced selection process in FL. To implement this selection strategy, the formulated integer programming problem is solved using standard optimization solvers, ensuring practical feasibility. The detailed implementation and solver configurations are further discussed in the \Cref{implementation}.

\subsection{Evaluation of Client Contributions}\label{evaluationclient}
Although reputation values help solve the client selection model \Cref{selection_model} for optimal decisions, client behavior can evolve over time due to various factors \citep{campos2022evaluating}. In real-world scenarios, clients may possess varying data quality and training capabilities, which directly impact their contributions to the global model. Moreover, clients may conceal their intentions and attempt to degrade the overall model quality in subsequent communication rounds through tactics such as Byzantine attack. Therefore, an accurate and fair evaluation mechanism is essential to ensure that reputation values evolve based on client contributions to the global model, effectively guiding future client selection.

To achieve this, we adopt the Shapley value from cooperative game theory, which systematically quantifies each participant’s contribution to a shared outcome \citep{shapley1953value}. Mathematically, the Shapley value for a client $C_i$ is defined as:
\begin{equation}\label{Shapley_value}
sv_i = \frac{1}{m} \sum_{\mathcal{T} \subseteq \mathbb{S} \setminus \{C_i\}} \frac{1}{\binom{m-1}{|\mathcal{T}|}} \left[ v(\mathcal{T} \cup \{C_i\}) - v(\mathcal{T}) \right],
\end{equation}
where $m$ is the total number of clients in the selected group $\mathbb{S}$, and $\mathcal{T}$ denotes any subset of participants excluding $C_i$. The function  $v(\mathcal{T})$ represents the utility or outcome (e.g., profit, accuracy, or any relevant metric) achieved by the subset $\mathcal{T}$. This formula quantifies each participant’s marginal contribution, ensuring fairness by considering all possible client combinations. 
The Shapley value has been widely applied across various domains, including data valuation \citep{ghorbani2019data}, where it quantifies the significance of individual data points in machine learning models. Its effectiveness in fairly assessing collaborative contributions makes it a natural choice for evaluating the impact of selected clients in FL. In the FL scenario, the Shapley value of client $C_i$ in $\mathbb{S}$ is formally defined in \Cref{Shapley_value_FL}.

\begin{definition}\label{Shapley_value_FL}
\textit{Shapley value in FL:} $SV_{\mathbb{S}^t} = CalculateSV(\mathbf{w}_{\mathbb{S}^t}, \mathcal{D}_{val})$. \textnormal{Here $\mathbf{w}_{\mathbb{S}^t}$ represents the model parameters uploaded in round $t$ by the selected clients in the group $\mathbb{S}^t$, and $\mathcal{D}_{val}$ is the validation dataset used to evaluate model performance. The Shapley value $sv_i^t$ for each client $C_i \in \mathbb{S}^t$, representing their individual contribution in round $t$, is expressed as:}
\begin{align}
  sv_i^t &= \frac{1}{m} \sum_{\mathcal{T} \subseteq \mathbb{S}^t \setminus \{C_i\}} 
    \frac{1}{\binom{m-1}{|\mathcal{T}|}} \Bigl[
    E\Bigl(\tfrac{1}{|\mathcal{T} \cup \{C_i\}|}
           \sum_{C_j \in \mathcal{T} \cup \{C_i\}}
           \mathbf{w}_j^t,\ 
           \mathcal{D}_{val}\Bigr) \nonumber\\
  & \quad 
    -\, E\Bigl(\tfrac{1}{|\mathcal{T}|}
               \sum_{C_j \in \mathcal{T}}
               \mathbf{w}_j^t,\ 
               \mathcal{D}_{val}\Bigr)
    \Bigr].
  \end{align}
\end{definition}

Here, $E(\mathbf{w}, \mathcal{D}_{val})$ evaluates model performance in the validation dataset $\mathcal{D}_{val}$, using the chosen metric (e.g., accuracy, area under the curve, or recall). The Shapley value quantifies each client's marginal contribution by comparing model performance with and without their participation. Specifically, it computes the difference $E(\mathcal{T} \cup {C_i}) - E(\mathcal{T})$, where $E(\mathcal{T})$ represents the performance of the model when trained with a subset $\mathcal{T}$ of clients, and $E(\mathcal{T} \cup {C_i})$ measures the performance when $C_i$ is additionally included. By aggregating these differences across all possible subsets $\mathcal{T} \subseteq \mathbb{S}^t \setminus {C_i}$, the Shapley value provides a fair assessment of individual contributions in each round.

Since a client's contribution to global model performance is directly influenced by the quality of their local dataset, the Shapley value naturally serves as a proxy for data quality. Clients with higher-quality data tend to provide more informative updates, leading to greater positive contributions, whereas lower-quality data may introduce noise, amplifying data quality decompensation and potentially degrading model performance. This property makes the Shapley value a reliable indicator of data quality, which forms the basis for reputation updates in \Cref{reputation_update}. In addition, we maintain a historical record \(\mathcal{SV}_{\text{his}} = \{\mathcal{SV}_1, \dots, \mathcal{SV}_n\}\), where $\mathcal{SV}_i$ stores the Shapley values of the client $C_i$ in the rounds they were selected. This historical record further supports the reputation update in \Cref{reputationupdate}, ensuring informed and adaptive adjustments.

Calculating Shapley values for all possible subsets has exponential complexity. However, practical approximations such as the Monte Carlo–Shapley method \citep{ghorbani2019data} and GTG-Shapley \citep{liu2022gtg} reduce computational overhead, making them suitable for large-scale FL. As cross-silo FL typically involves a limited number of participants, we employ exact Shapley value computation to ensure a precise contribution evaluation.

\subsection{Reputation Update}\label{reputationupdate}
The Shapley value provides a quantitative measure of client contributions, which forms a key component in reputation updates. However, effective client selection must balance both performance and economic feasibility. To achieve this, reputation updates incorporate bid prices alongside Shapley values, ensuring that high-quality participants are incentivized while maintaining cost-effectiveness. Specifically, positively contributing clients ($sv^t_i > 0$) receive increased reputation, while negative or near-zero values ($sv^t_i \leq 0$) trigger penalization. The update rule is defined as follows:
\begin{definition}\label{reputation_update}
    \textit{Reputation Update Rule.} 
    \textnormal{Reputation updates depend on each client’s Shapley value $sv_i^t$, bid price $B_i$, and recent performance history $err_i$. First, we define: \begin{equation}
        S_{pos} = \sum_{C_i \in \mathbb{S}^t, sv_i^t > 0} sv_i, \quad B_{pos} = \sum_{C_i \in \mathbb{S}^t, sv_i^t > 0} B_i
    \end{equation}
    where $S_{pos}$ is the total Shapley value of all positively contributing clients, and $B_{pos}$ is the total bid price of those clients. The reputation update formula is expressed as follows:}
    \begin{equation}
    R_i = \begin{cases}
    R_i - \psi \cdot \rho^{err_i} & \text{if } C_i \in \mathbb{S}^t \text{ and } sv_i^t \leq 0, \\
    R_i + \omega \cdot \left( 1 - \exp\left(-\frac{sv_i^t / S_{pos}}{B_i / B_{pos}}\right) \right) & \text{if } C_i \in \mathbb{S}^t \text{ and } sv_i^t > 0\\
    R_i & \text{if } C_i \notin \mathbb{S}^t.
    \end{cases}
    \end{equation}
\end{definition}

\begin{algorithm}[htbp]
    \caption{Shapley-Bid Reputation Optimized Client Selection}
    \label{alg:dynamic_client_selection}
    \begin{algorithmic}[1]
    \STATE \textbf{Initialize:} Model parameters $\mathbf{w}^0$, Reputation $\mathcal{R} = \{R_1, \dots, R_n\}$, bid price $\mathcal{B} = \{B_1, \dots, B_n\}$
    
    \FOR {round $t = 1$ to $T$}
        \STATE $R_{\text{th}} \leftarrow \frac{1}{n}\sum_{i=1}^{n} R_i$
        \FOR{each $C_i \in \mathcal{C}$}
            \STATE $z(R_i) \leftarrow 
            \begin{cases} 
                \gamma (R_{\text{th}} - R_i)^\beta, & \text{if } R_i \leq R_{\text{th}} \\
                (R_i - R_{\text{th}})^\alpha, & \text{if } R_i > R_{\text{th}} 
            \end{cases}$
        \ENDFOR
        \STATE $\mathbb{S}^t \leftarrow Select(\mathcal{Z}, \mathcal{B}, B_{\text{budget}}, \mathcal{H}^t)$
        \STATE Distribute global model $\mathbf{w}^t$ to clients in $\mathbb{S}^t$
        \FOR{each $C_i \in \mathbb{S}^t$}
            \STATE $\mathbf{w}_i^{t} \leftarrow \mathbf{w}^{t-1} - \eta \mathbf{g}_i^{t-1}$
        \ENDFOR
        \STATE $\mathbf{w}^{t} \leftarrow \text{Aggregate}(\mathbf{w}_{\mathbb{S}^t}^{t})$
        \STATE $SV_{\mathbb{S}^t} \leftarrow \text{CalculateSV}(\mathbf{w}_{\mathbb{S}^t}^{t}, \mathcal{D}_{val})$
        \STATE $S_{pos} \leftarrow \sum_{C_i \in \mathbb{S}^t, sv_i^t > 0} sv_i^t$
        \STATE $B_{pos} \leftarrow \sum_{C_i \in \mathbb{S}^t, sv_i^t > 0} B_i$
        \FOR{each $C_i \in \mathbb{S}^t$}
            \STATE $R_i \leftarrow 
            \begin{cases} 
                R_i - \psi \cdot \rho^{err_i}, & \text{if } sv_i^t \leq 0 \\
                R_i + \omega \cdot \left( 1 - \exp\left(-\frac{sv_i^t / S_{pos}}{B_i / B_{pos}}\right) \right) & \text{if } sv_i^t > 0
            \end{cases}$
        \ENDFOR
        \STATE \textbf{Update} $\mathcal{SV}_{\text{his}}, \mathcal{H}^t$     
    \ENDFOR
    \end{algorithmic}
\end{algorithm}
Here, $\omega$ and $\psi$ represent the reward and punishment coefficients, respectively, while $err_i$ tracks the number of times client $C_i$ has demonstrated poor performance (i.e., $sv_i^t \leq 0$) in the last five rounds they were selected. The set $\mathcal{SV}_{\text{his}}$, which stores each client’s Shapley values across rounds, is used to calculate $err_i$ by identifying rounds with negative contributions. The penalty factor $\rho$ increases exponentially with repeated poor performance, ensuring harsher penalties for clients who consistently underperform.

The penalty mechanism swiftly filters out clients whose updates degrade model performance, whether due to unreliable data or overfitting. In contrast, positively contributing clients receive reputation updates based on both their impact and bid price, ensuring that high-quality, cost-effective participants are incentivized for future selection. This approach encourages clients to submit realistic bids while maintaining high-quality selections in subsequent rounds. Additionally, the use of an exponential reward function prevents excessively rapid reputation increases, mitigating the risk of overfitting the global model to specific clients. For clients not selected in a given round, their reputation remains unchanged, ensuring fairness across rounds.

Thus, SBRO-FL forms a closed-loop framework, integrating bidding, reputation-based selection, contribution evaluation, and adaptive reputation updates. This iterative process continuously refines client participation, ensuring a balance between data reliability, budget constraints, and incentive alignment. The full procedure is detailed in Algorithm \ref{alg:dynamic_client_selection}.

\section{Experiments Setup}\label{experimentsetup}
This section presents the experimental setup used to evaluate the proposed SBRO-FL method. \Cref{datasets} outlines the datasets used and describes how we simulate data quality decompensation via label flipping. \Cref{baselines} introduces the baseline methods for comparison, and \Cref{implementation} provides the implementation details, including the FL framework, model architectures, and the parameters used to simulate the FL environment under budget constraints.

\subsection{Data Partitioning}\label{datasets}
We utilized four widely recognized benchmark datasets—FashionMNIST \citep{xiao2017fashion}, EMNIST-Letter \citep{cohen2017emnist}, SVHN \citep{netzer2011reading}, and CIFAR-10 \citep{krizhevsky2009learning}—for image classification tasks in a FL environment. These datasets were selected for their diverse characteristics, including variations in image resolution, color channels, and class distributions, providing a comprehensive basis for evaluating FL algorithms under different conditions. \Cref{tab:dataset_stats} summarizes the key statistics of 
each dataset, including the number of classes and train/test splits.
\begin{table}[htbp]
  \captionsetup{
      font=small,
      labelfont=bf,
      labelsep=none,
      justification=raggedright,
      singlelinecheck=false
  }
  \setlength{\abovecaptionskip}{0pt} 
  \caption{\\
  Statistics of the datasets used in FL experiments}
  \label{tab:dataset_stats}
  \centering
  \footnotesize
  \begin{tabularx}{\linewidth}{
    >{\raggedright\arraybackslash}p{2cm}
    >{\centering\arraybackslash}X
    >{\centering\arraybackslash}X
    >{\centering\arraybackslash}X
    >{\centering\arraybackslash}X}
  \toprule
  Dataset         & Total Samples & Training Samples & Test Samples & Classes \\
  \midrule
  EMNIST-letter   & 124,800       & 88,800           & 36,000       & 26 \\
  FashionMNIST    & 70,000        & 60,000           & 10,000       & 10 \\
  SVHN            & 600,000       & 73,257           & 53,608       & 10 \\
  CIFAR-10        & 60,000        & 50,000           & 10,000       & 10 \\
  \bottomrule
  \end{tabularx}
  \end{table}

To simulate a realistic cross-silo FL scenario, the training data for each dataset was limited to 10,000 samples, which were randomly partitioned among 40 clients. Each client was assigned an equal portion of the training data, and in each FL round, a subset of clients was selected for training based on the proposed selection model. The central server aggregated client updates to refine the global model over multiple communication rounds.

To examine the impact of data quality decompensation on model performance, label flipping was applied to 32 of the clients. These clients were randomly divided into four groups of eight, with label flipping proportions set at 90\%, 80\%, 70\% and 60\%, respectively. The remaining eight clients retained their original labels and served as high-quality participants.

Label flipping is chosen as a straightforward yet practical method to simulate data quality decompensation. Compared to other noise models, label flipping more directly affects the training objective and mimics malicious and unintentional mislabeling scenarios. By manipulating the flipping ratios, it is possible to assess the effectiveness of the method in identifying and mitigating the impact of low-quality updates.

\subsection{Baselines}\label{baselines}
To evaluate the effectiveness of our proposed SBRO-FL approach, we compare it with three baseline methods that capture a broad spectrum of client selection strategies:
\begin{itemize}[leftmargin=*, labelsep=0.5em, nosep]
    \item \textbf{Random Selection (RS-FL)}: In each round, the server randomly chooses a subset of clients within the budget constraint \citep{mcmahan2017communication}. This method does not consider client reliability or historical performance and serves as a baseline to assess whether reputation-based selection improves FL outcomes.
    \item \textbf{High-Quality Random Selection (HQRS-FL)}: Clients are randomly selected from a high-quality subset (unaffected by label flipping) while maintaining the same budget constraint. This represents an ``oracle'' scenario where the server selects only high-quality clients based on prior knowledge of their data quality. Although such an assumption is unrealistic in practical FL settings, HQRS-FL serves as a reference to assess how well SBRO-FL approximates high-quality selection without explicit quality labels.
    \item \textbf{All Clients Selected without Budget (All-FL)}: All clients, regardless of their data quality or bid price, are selected in each round without budget constraints. However, in real-world FL deployments, unrestricted participation often leads to lower model robustness, as unreliable updates from low-quality clients degrade overall performance. This comparison allows us to evaluate whether SBRO-FL effectively balances inclusiveness and reliability within a realistic budget-constrained setting.
\end{itemize}

Although there are many advanced FL methods, such as those that focus on incentive mechanisms, communication efficiency, or adversarial robustness, these approaches typically optimize a single objective. In contrast, our work addresses data quality decompensation, incentive compatibility, and budget constraints simultaneously. A direct comparison with single-focus methods would not fully capture the effectiveness of our integrated approach. Therefore, we adopt these three baselines, which collectively benchmark the core aspects of client selection in our setting.

\subsection{Implementation Details}\label{implementation}
The experiments were carried out using the FLEXible platform \citep{herrera2024flexflexiblefederatedlearning}, an open source framework that provides a comprehensive set of tools for deep learning and machine learning in federated environments. FLEXible allows full customization of the FL scenario, from foundational components to high-level configurations, making it well suited to evaluate the proposed method.

CNN architectures were adapted to the complexity of the data et: two convolutional layers for FashionMNIST and EMNIST, and three for CIFAR-10 and SVHN. All models were trained with mini-batch SGD (batch size = 16) with an initial learning rate of $\eta = 0.01$. Client updates were aggregated using FedAvg over 300 communication rounds.

The bid prices of the clients were generated following a Gaussian distribution ($\mu = 10$, $\sigma^2 = 1$) to simulate natural variation in the valuation of the clients' data and computational costs, and the task budget was fixed at $B_{\text{budget}} = 45$. The selection process incorporated a decay factor $\delta = 0.5$ to balance participation diversity. The prospect theory parameters were established as $\alpha = 0.15$, $\beta = 0.3$, and $\gamma = 1$, after prior studies and preliminary adjustments. The client selection problem was solved using the PuLP linear programming solver. To improve computational stability and ensure effective selection, standard preprocessing techniques were applied before solving.

The complete implementation and experimental configurations are available on GitHub\footnote{GitHub repository: \url{https://github.com/ari-dasci/S-SBRO-FL}}, ensuring full reproducibility.
\section{Experimental Results}\label{experimentalresults}
This section presents the experimental results of SBRO-FL compared with the baseline methods across four benchmark datasets. We evaluated (i) the final performance and training stability of the global model across multiple datasets (see \Cref{PSE}), and (ii) the robustness of SBRO-FL against low-cost interference (see \Cref{RALC}).

\subsection{Performance and Stability Evaluation}\label{PSE}
\begin{figure*}[t]
\centering
\includegraphics[width = \textwidth]{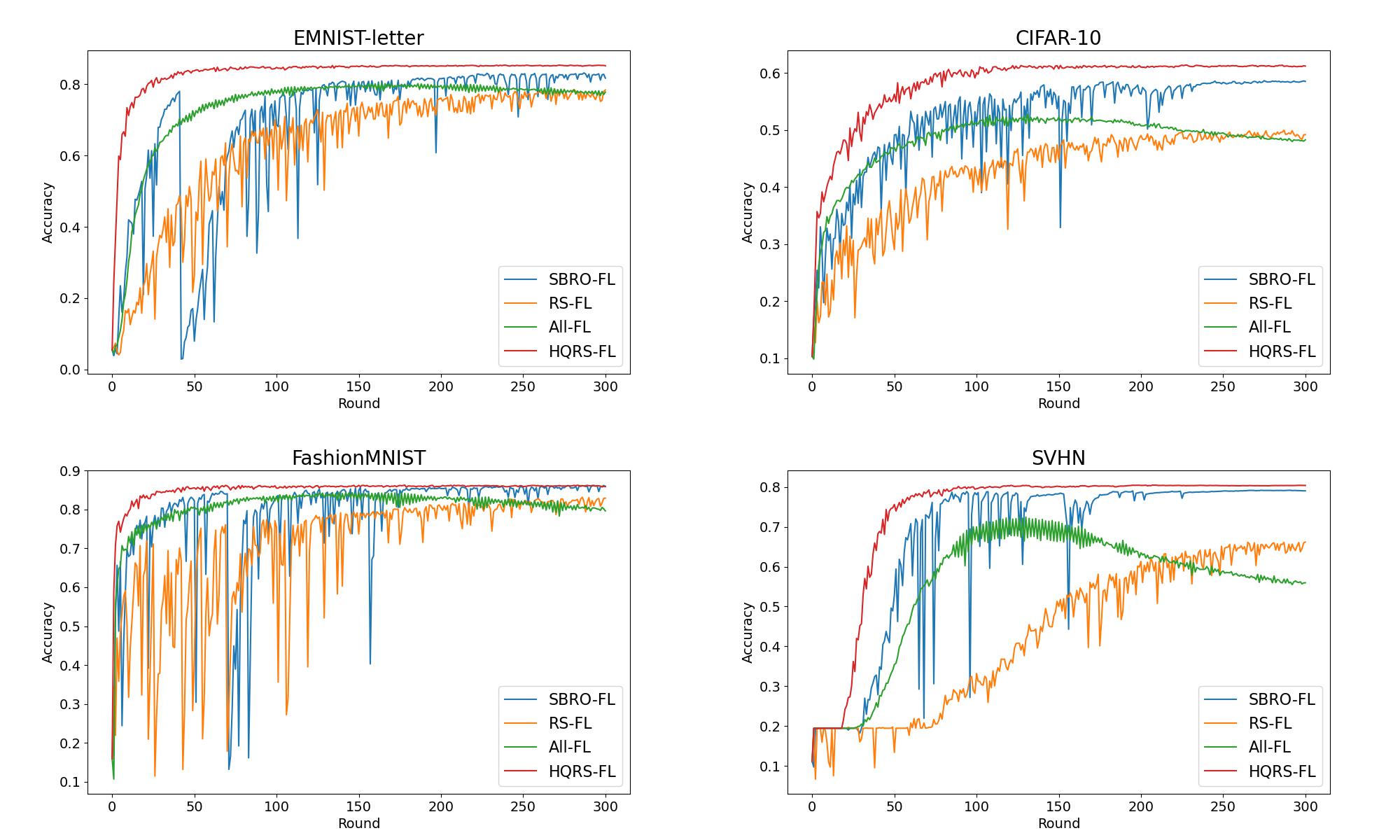}
\caption{Trends in Model Accuracy: Comparing SBRO-FL and Baseline Methods Across Diverse Datasets.}
\label{Model Accuracy Curves}
\end{figure*}

In this subsection, we first compare the accuracy of the global model in the final round of each method in \Cref{tab:model_accuracy1}, then examine their convergence stability by plotting accuracy trends over multiple communication rounds (see \Cref{Model Accuracy Curves}).

\begin{table*}[htbp]
    \footnotesize
    \captionsetup{
      font=small,
      labelfont=bf,
      labelsep=none,
      justification=raggedright,
      singlelinecheck=false
    }
    \setlength{\abovecaptionskip}{0pt} 
    \caption{\\Final round global model accuracy of different methods on four datasets. The variance across the final training rounds is negligible (less than $10^{-4}$), indicating that observed accuracy improvements are not due to random fluctuations. Note: ``Gain" represents the accuracy improvement of SBRO-FL over RS-FL, calculated as $(\text{SBRO-FL} - \text{RS-FL}) / \text{RS-FL} \times 100\%$.}
    \label{tab:model_accuracy1}
    \centering
    \begin{tabularx}{\linewidth}{p{2.5cm} X X X X p{2.5cm}}
    \toprule
    Dataset         & SBRO-FL       & RS-FL          & HQRS-FL        & All-FL       & Gain (\% over RS-FL) \\
    \midrule
    EMNIST-letter   & 0.8166 & 0.7835         & \textbf{0.8515}         & 0.7777       & +4.2\% \\
    FashionMNIST    & \textbf{0.8600} & 0.8294         & 0.8590         & 0.7966       & +3.7\% \\
    CIFAR-10        & 0.5853 & 0.4920         & \textbf{0.6121}         & 0.4828       & +19.0\% \\
    SVHN            & 0.7902 & 0.6617         & \textbf{0.8036}         & 0.5595       & +19.4\% \\
    \midrule
    Average         & 0.7630 & 0.6917         & \textbf{0.7816}         & 0.6541       & +10.3\% \\
    \bottomrule
    \end{tabularx}
\end{table*}

\Cref{tab:model_accuracy1} reports the accuracy of the global model in the final round for different methods. SBRO-FL consistently outperforms RS-FL in all datasets, demonstrating its ability to prioritize high-quality client updates while operating under budget constraints. Furthermore, we computed the variance of the final 20 rounds for each method, which remained negligibly small. This confirms that random fluctuations did not influence the analysis, reinforcing the robustness of the reported results.

In particular, CIFAR-10 and SVHN exhibit larger gains around 19\%, probably due to their greater complexity and higher susceptibility to label-flipping noise. In contrast, simpler datasets such as FashionMNIST show smaller relative gains, as even random selection can achieve a relatively high baseline accuracy.

Another interesting observation from the FashionMNIST experiment is that SBRO-FL achieved a slightly higher final accuracy than HQRS-FL, the ideal ``oracle" scenario. This phenomenon can be attributed to the ability of SBRO-FL to learn additional knowledge from clients with label noise, while HQRS-FL exclusively selects clients from a pool entirely free of label noise. This finding underscores the importance of diversity in client selection. 

Additionally, although All-FL involves all available clients, it does not consistently improve accuracy and incurs higher computational costs. This highlights the need for strategic client selection, as indiscriminate participation amplifies data quality decompensation, leading to accumulated unreliable updates and global performance degradation.

\Cref{Model Accuracy Curves} provides a visual comparison of the accuracy of the global model in all training rounds, providing insight into the stability and convergence behavior of different methods. The trends observed here reinforce the findings from \Cref{tab:model_accuracy1}, demonstrating the ability of SBRO-FL to adaptively optimize client selection over time. During training, SBRO-FL exhibits slight fluctuations and initially lags behind RS-FL. This is due to the ongoing assessment of client reliability, during which some lower-quality clients may still be selected. However, as training progresses, SBRO-FL progressively refines its selection, leading to a sustained improvement in both accuracy and stability. In later rounds, high-quality clients have been consistently prioritized, enabling SBRO-FL to surpass RS-FL and maintain more stable performance. In contrast, All-FL exhibits a ``peak-and-drop" behavior, where an initial increase in accuracy is followed by a decline due to the incorporation of noisy updates, ultimately degrading overall performance. Although All-FL benefits from a higher number of participants, the lack of selective filtering leads to poor long-term generalization.

For more complex datasets such as CIFAR-10 and SVHN, the advantage of SBRO-FL becomes even more pronounced. In these cases, SBRO-FL not only achieves higher accuracy, but also demonstrates faster and more stable convergence compared to RS-FL. These trends confirm that SBRO-FL’s selection mechanism successfully adapts over time, ensuring long-term stability and performance improvements.

\subsection{Robustness Against Low-Cost Interference}\label{RALC}
FL systems face the risk of adversarial bidding, where unreliable clients reduce their bids to increase the probability of selection. This section evaluates SBRO-FL’s resilience to such low-cost interference, testing whether it can effectively prioritize high-quality clients despite financial manipulations.

To model a realistic low-bid interference scenario, bid prices are assigned based on label-flipping ratios: clients with flipping rates of 90\%, 80\%, 70\%, 60\%, and 0\% receive bids of 6, 8, 10, 12, and 14, respectively. This setup mimics a practical challenge in which low-quality clients strategically lower their bids to increase selection chances.

\begin{table*}[htbp]
    \footnotesize
    \captionsetup{
      font=small,
      labelfont=bf,
      labelsep=none,
      justification=raggedright,
      singlelinecheck=false
    }
    \setlength{\abovecaptionskip}{0pt} 
    \caption{\\ Final round global model accuracy of different methods on four datasets after adjusting the bid strategy. ``Gain" represents the improvement of SBRO-FL over RS-FL, calculated as $(\text{SBRO-FL} - \text{RS-FL}) / \text{RS-FL} \times 100\%$.}
    \label{tab:model_accuracy_price}
    \centering
    \begin{tabularx}{\linewidth}{p{2.5cm} X X X X p{2.5cm}} 
    \toprule
    Dataset         & SBRO-FL       & RS-FL          & HQRS-FL        & All-FL       & Gain (\% over RS-FL) \\
    \midrule
    EMNIST-letter   & 0.8176        & 0.7735         & \textbf{0.8532} & 0.7605       & +5.7\% \\
    FashionMNIST    & 0.8499        & 0.7638         & \textbf{0.8590} & 0.8010       & +11.3\% \\
    CIFAR-10        & 0.5999        & 0.4818         & \textbf{0.6073} & 0.4817       & +24.5\% \\
    SVHN            & 0.7808        & 0.5935         & \textbf{0.8055} & 0.5590       & +31.6\% \\
    \midrule
    Average         & 0.7620        & 0.6532         & \textbf{0.7812} & 0.6506       & +16.7\% \\
    \bottomrule
    \end{tabularx}
\end{table*}
    
As shown in \Cref{tab:model_accuracy_price}, SBRO-FL effectively selects high-quality clients, even amidst varying data quality and low-bid interference. SBRO-FL significantly outperforms both RS-FL and All-FL across all datasets, with particularly strong results on the more complex SVHN and CIFAR-10 datasets. On average, SBRO-FL achieves an improvement in accuracy of 10.89\% over RS-FL and an improvement of 1.92\% over All-FL, underscoring its robustness in client selection despite data variability and low bids.

\begin{figure*}[t]
\centering
\includegraphics[width = \textwidth]{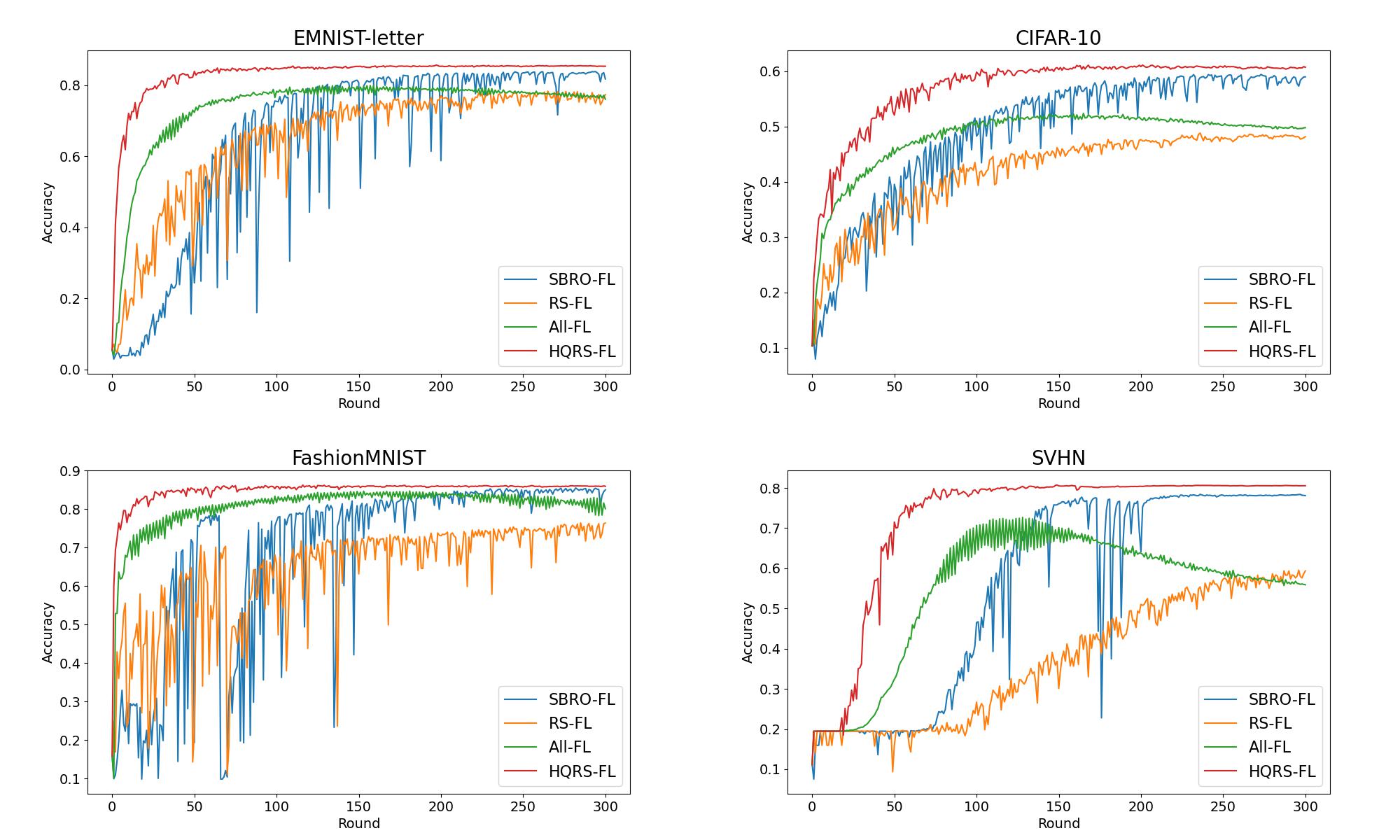}
\caption{Trends in Model Accuracy: Evaluating SBRO-FL and Baseline Methods Under Low-Cost Interference Across Datasets.}
\label{Model Accuracy Curves2}
\end{figure*}
  
\Cref{Model Accuracy Curves2} shows the accuracy of the global model per round for the four methods after adjusting the bid strategy. This price-sensitive experiment reveals a key insight: The SBRO-FL decision-making mechanism, which balances reputation and bid price, is resistant to manipulation by low bids. Even when lower bids are present, SBRO-FL successfully avoids selecting clients with poor data quality, highlighting the method’s emphasis on data quality over short-term financial incentives to maintain high model performance.

When comparing these results with \Cref{Model Accuracy Curves}, we observe similar trends: SBRO-FL consistently outperforms RS-FL and All-FL in stability and convergence speed, even with low-bid interference. Although SBRO-FL shows slight oscillations in accuracy, these variations are minimal, indicating that the low-bid strategy has a limited impact on its performance. In general, SBRO-FL maintains its advantage, reaffirming its robustness under challenging conditions.
    
Interestingly, a comparison between \Cref{Model Accuracy Curves} and \Cref{Model Accuracy Curves2} shows that while All-FL occasionally converges faster than SBRO-FL, it eventually experiences a decline in accuracy due to the continual incorporation of noisy updates. This underscores the importance of excluding low-quality clients in FL. The negative impact of such clients, misaligned updates and poor data quality, can lead to persistent degradation in global model performance. Consequently, SBRO-FL's approach to filtering out low-quality clients proves essential to achieve both stability and superior final performance.

\subsection{Discussion on Component Contributions}\label{discussion}
While the experimental results demonstrate the effectiveness of SBRO-FL as a unified framework, a detailed ablation study remains a valuable direction for further investigation. Specifically, evaluating the isolated impact of each component, such as the bidding mechanism without reputation modeling, the reputation system without Shapley-based evaluation, or the use of simpler heuristics in place of Shapley values, would help quantify their respective contributions to overall performance. Such an analysis would provide deeper insights into the role of each module in improving accuracy, robustness, and cost-efficiency, and would further justify the design choices made in SBRO-FL. We consider this to be an important avenue for future empirical work.

\section{Conclusion and future work}\label{conclusion}
This work presents SBRO-FL, a unified client selection framework for FL in silos that addresses the intertwined challenges of data quality decompensation, incentive compatibility, and budget constraints. By combining a reputation-driven bidding mechanism with cost-aware optimization, SBRO-FL ensures that client selection reflects both historical contributions and economic feasibility. The integration of a Shapley value-based contribution evaluation and a prospect-theory-inspired reputation update enables robust and adaptive client participation over time. Our extensive empirical evaluation in four benchmark datasets shows that SBRO-FL consistently outperforms traditional random and inclusive selection strategies, even in the presence of adversarial bidding and noisy data. These results confirm the practical effectiveness of our approach in enhancing the robustness and efficiency of FL systems.

\subsection{Limitations}
While SBRO-FL demonstrates strong performance across diverse datasets and adversarial scenarios, certain limitations remain. The exact computation of Shapley values, while offering fair and precise contribution assessments, incurs combinatorial complexity that may not scale efficiently to federations involving hundreds or thousands of clients. Although this is less critical in cross-silo settings where the number of clients is typically small, future work should investigate scalable approximations or surrogate contribution metrics to maintain fairness and computational efficiency in larger deployments. Additionally, our current implementation assumes honest bid submissions; future versions could incorporate mechanisms for bid verification or robustness against strategic misreporting.

From a computational standpoint, the client selection problem formulated in SBRO-FL is a variant of the 0–1 knapsack problem, which is known to be NP hard. This implies that finding an optimal solution becomes computationally intensive as the number of clients increases, necessitating efficient solvers or approximations in large-scale deployments. Regarding convergence, SBRO-FL builds upon standard FL frameworks like FedAvg, whose convergence under non-i.i.d. settings has been established in prior work. Since our selection mechanism preserves the core iterative structure and does not alter local training dynamics, it inherits similar convergence behavior under bounded-variance assumptions. Furthermore, although exact Shapley value computation is used in this study, approximate methods such as Monte Carlo sampling and GTG-Shapley provide theoretical error bounds and can be adopted in future work to ensure scalability while retaining fairness.

\subsection{Future Work}
While SBRO-FL demonstrates strong performance in controlled experimental settings, several promising directions remain for future research. First, real-world deployment studies in industrial or healthcare cross-silo environments could validate the method's robustness under real operational constraints. Second, exploring adaptive or learnable reputation metrics—possibly using neural attention mechanisms or metalearning—may enhance the selection dynamics beyond fixed prospect-theoretic functions. Finally, scaling SBRO-FL to larger federations with hundreds or thousands of clients will require more computationally efficient approximations of Shapley values, such as Monte Carlo or GTG-based methods, to maintain scalability without compromising fairness or accuracy.

\section*{Acknowledgements}
This work was supported in part by the National Natural Science Foundation of China under Grant 72171065 and in part by the Open Fund of Shaanxi Key Laboratory of Information Communication Network and Security under Grant ICNS201807. 

It also results from the Strategic Project IAFER-Cib (C074/23), as a result of the collaboration agreement signed between the National Institute of Cybersecurity (INCIBE) and the University of Granada. This initiative is carried out within the framework of the Recovery, Transformation and Resilience Plan funds, financed by the European Union (Next Generation). 

Qinjun Fei acknowledges the support of the China Scholarship Council program (Project ID: 202308330099).





\bibliographystyle{plainnat}
\bibliography{mybibfile}
\end{document}